\documentclass{article}

\usepackage[preprint]{neurips_2020}

\usepackage[utf8]{inputenc} 
\usepackage[T1]{fontenc}    
\usepackage{hyperref}       
\usepackage{url}            
\usepackage{booktabs}       
\usepackage{amsfonts}       
\usepackage{nicefrac}       
\usepackage{microtype}      

\usepackage{subfigure}
\usepackage{float}

\usepackage{wrapfig}
\usepackage{xcolor}
\usepackage{amsmath}
\usepackage{graphicx, caption}
\usepackage{mathtools}
\usepackage{algorithm}
\usepackage{algpseudocode}
\usepackage{cleveref}

\DeclarePairedDelimiterX{\infdivx}[2]{(}{)}{%
  #1\;\delimsize\|\;#2%
}

\definecolor{dark-green}{RGB}{0, 153, 51}
\definecolor{dark-blue}{RGB}{0, 0, 153}
\definecolor{dark-red}{RGB}{153, 0, 0}

\title{Safety Aware Reinforcement Learning (SARL)}

\author{
  Santiago Miret    \\
  Intel AI\\
  \texttt{santiago.miret@intel.com} \\
    \And
  Somdeb Majumdar \\
  Intel AI \\
  \texttt{somdeb.majumdar@intel.com} \\
  \And
  Carroll Wainwright \\
  Partnership on AI \\
  \texttt{carroll@partnershiponai.org} \\
}

\begin{document}

\maketitle
\begin{abstract}

As reinforcement learning agents become increasingly integrated into complex, real-world environments, designing for safety becomes a critical consideration. We specifically focus on researching scenarios where agents can cause undesired side effects while executing a policy on a primary task. Since one can define multiple tasks for a given environment dynamics, there are two important challenges. First, we need to abstract the concept of safety that applies broadly to that environment independent of the specific task being executed. Second, we need a mechanism for the abstracted notion of safety to modulate the actions of agents executing different policies to minimize their side-effects. In this work, we propose Safety Aware Reinforcement Learning (SARL) -- a framework where a virtual safe agent modulates the actions of a main reward-based agent to minimize side effects. The safe agent learns a task-independent notion of safety for a given environment. The main agent is then trained with a regularization loss given by the distance between the native action probabilities of the two agents. Since the safe agent effectively abstracts a task-independent notion of safety via its action probabilities, it can be ported to modulate multiple policies solving different tasks within the given environment without further training. We contrast this with solutions that rely on task-specific regularization metrics and test our framework on the SafeLife Suite, based on Conway's Game of Life, comprising a number of complex tasks in dynamic environments. We show that our solution is able to match the performance of solutions that rely on task-specific side-effect penalties on both the primary and safety objectives while additionally providing the benefit of generalizability and portability.

\end{abstract}

\section{Introduction} \label{introduction}
Reinforcement learning (RL) algorithms have seen great research advances in recent years, both in theory and in their applications to concrete engineering problems. The application of RL algorithms extends to computer games \citep{mnih2013playing, silver2017mastering}, robotics \citep{gu2017deep} and recently real-world engineering problems, such as microgrid optimization \citep{liu2018distributed} and hardware design \citep{mirhoseini2020chip}. As RL agents become increasingly prevalent in complex real-world applications, the notion of safety becomes increasingly important. Thus, safety related research in RL has also seen a significant surge in recent years \citep{Zhang2020, Brown2020, Mell2019, Cheng2019, Rahaman2020}.

\subsection{Side Effects in Reinforcement Learning Environments}
Our work focuses specifically on the problem of side effects, identified as a key topic in the area of safety in AI by  \citet{amodei2016safetyproblems}. Here, an agent's actions to perform a task in its environment may cause undesired, and sometimes irreversible, changes in the environment. A major issue with measuring and investigating side effects is that it is challenging to define an appropriate side-effect metric, especially in a general fashion that can apply to many settings. The difficulty of quantifying side effects distinguishes this problem from safe exploration and traditional motion planning approaches that focus primarily on avoiding obstacles or a clearly defined failure state \citep{amodei2016safetyproblems, Zhu2020}. As such, when learning a task in an unknown environment with complex dynamics, it is challenging to formulate an appropriate environment framework to jointly encapsulate the primary task and side effect problem.

Previous work on formulating a more precise definition of side effects includes work by \citet{turner2019aup} on conservative utility preservation and by \citet{krakovna2018penalizing} on relative reachability. These works investigated more abstract notions of measuring side effects based on an analysis of changes, reversible and irreversible, in the state space itself. While those works have made great progress on advancing towards a greater understanding of side effects, they have generally been limited to simple grid world environments where the RL problem can often be solved in a tabular way and value function estimations are often not prohibitively demanding. Our work focuses on expanding the concept of side effects to more complex environments, generated by the SafeLife suite \citep{Wainwright2020}, which provides more complex environment dynamics and tasks that cannot be solved in a tabular fashion. \citet{turner2020avoiding} recently extended their approach to environments in the SafeLife suite, suggesting that attainable utility preservation can be used as an alternative to the SafeLife side metric described in \citet{Wainwright2020} and \Cref{safelife-env}. The primary differentiating feature of SARL is that it is metric agnostic, for both the reward and side effect measure, making it orthogonal and complimentary to the work by \citet{turner2020avoiding}.

In this paper, we make the following contributions which, to the best of our knowledge, are novel additions to the growing field of research in RL safety:
\begin{itemize}
    \item \emph{SARL}: a flexible, metric agnostic RL framework that can modulate the actions of a trained RL agent to trade off between task performance and a safety objective. We utilize the distance between the action probability distributions of two policies as a form of regularization during training.
    \item A generalizeable notion of safety that allows us to train a safe agent independent of specific tasks in an environment and port it across multiple complex tasks in that environment. 
    
\end{itemize}
We provide a description of the SafeLife suite in Section \ref{safelife-env}, a detailed description of our method in Section \ref{method}, our experiments and results for various environments in \Cref{experiments} and \Cref{results} respectively, as well as a discussion in Section \ref{discussion}.
\newcommand{\inlinesprite}[1]{
  \begingroup\normalfont\hspace{-0.5em}
  \raisebox{-.1\height}{\includegraphics[height=0.65em, clip, trim={#1}]{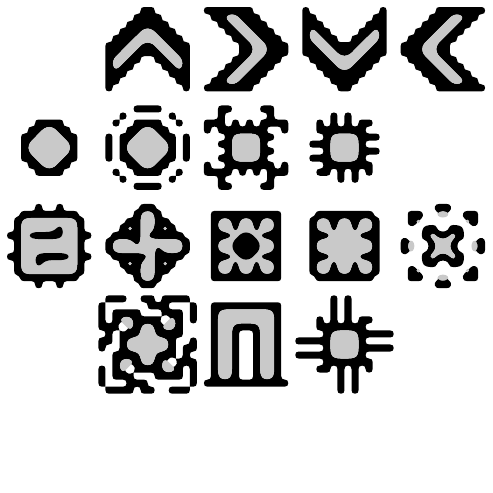}}
  \hspace{-0.22em}\endgroup
}
\DeclareRobustCommand {\agentsprite}    {\inlinesprite{2cm 4cm 2cm 0cm}}
\DeclareRobustCommand {\lifesprite}     {\inlinesprite{0cm 3cm 4cm 1cm}}
\DeclareRobustCommand {\hardlifesprite} {\inlinesprite{1cm 3cm 3cm 1cm}}
\DeclareRobustCommand {\weedsprite}     {\inlinesprite{2cm 3cm 2cm 1cm}}
\DeclareRobustCommand {\plantsprite}    {\inlinesprite{3cm 3cm 1cm 1cm}}
\DeclareRobustCommand {\treesprite}     {\inlinesprite{3cm 1cm 1cm 3cm}}
\DeclareRobustCommand {\icesprite}      {\inlinesprite{0.02cm 2.02cm 4cm 2cm}}
\DeclareRobustCommand {\fountainsprite} {\inlinesprite{1cm 2cm 3cm 2cm}}
\DeclareRobustCommand {\wallsprite}     {\inlinesprite{2cm 2cm 2cm 2cm}}
\DeclareRobustCommand {\cratesprite}    {\inlinesprite{3cm 2cm 1cm 2cm}}
\DeclareRobustCommand {\predsprite}     {\inlinesprite{4cm 2cm 0cm 2cm}}
\DeclareRobustCommand {\spawnsprite}    {\inlinesprite{1cm 1cm 3cm 3cm}}
\DeclareRobustCommand {\exitsprite}     {\inlinesprite{2cm 1cm 2cm 3cm}}

\section{The SafeLife Environment} \label{safelife-env}

\begin{figure}[h]
    \centering
    \includegraphics[width=1.6in]{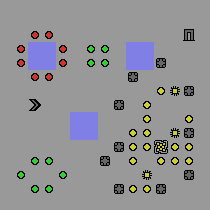}
    \caption{A simple level of the \emph{SafeLife} environment containing an agent~\mbox{(\agentsprite)}, a spawner~\mbox{(\spawnsprite)}, crates~\mbox{(\cratesprite)}, and cells of life. The agent's goal is to remove unwanted red cells (\emph{prune task}) and to create new patterns of life in the blue squares (\emph{append task}). Once the agent has satisfactorily completed its goals it can leave via the level exit~\mbox{(\exitsprite)}.
    Note that all level boundaries wrap; they have toroidal topology.}
    \label{fig:simple-spawner}
\end{figure}

The SafeLife suite \citep{Wainwright2020} creates complex environments of systems of cellular automata based on a set of rules from Conway's Game of Life \citep{gardner1970fantastic} that govern the interactions between, and the state (alive or dead) of, different cells:
\begin{itemize}
    \item any dead cell with exactly three living neighbors becomes alive;
    \item any live cell with less than two or more than three neighbors dies (as if by under- or overpopulation); and
    \item every other cell retains its prior state.
\end{itemize}
In addition to the basic rules, SafeLife enables the creation of complex, procedurally generated environments through special cells, such as a spawner that can create new cells and dynamically generated patterns. The agent can generally perform three tasks: \emph{navigation}, \emph{prune} and \emph{append} which are illustrated in Figure \ref{fig:simple-spawner} taken from \citet{Wainwright2020}.

The flexibility of SafeLife enables the creation of still environments, where the cell patterns do not change over time without agent interference, and dynamic environments, where the cell patterns do change over time without agent interference. The dynamic environments create an additional layer of difficulty, as the agent now needs to learn to distinguish between variations in the environment that are triggered by its own actions versus those that are caused by the dynamic rules independent of its actions. As described in Section \ref{experiments}, our experiments focus on the prune and append tasks in still and dynamic environments: \emph{prune-still}, \emph{prune-dynamic}, \emph{append-still}, \emph{append-dynamic}.

\subsection{SafeLife Side Effect Metric}
The Safelife suite calculates the overall side effects at the end of the episode by taking a time-average of the state of the environment for a series of steps after the episode ends. This process is meant to ensure that the dynamics of the environment stabilize after the end of the episode. Stabilization is particularly important for dynamic environments, where the inherent variations in the environment can amplify effects many timesteps beyond the end of the episode. 
In addition to the overall side effect at the end of the episode, SafeLife also has the option of producing an impact penalty for each environment step during agent training. In the original SafeLife paper, \citet{Wainwright2020} use that impact penalty to change the agent's reward function directly, whereas we are using the side effect information channel to train a virtual safety agent that generalizes across tasks and environment settings. The results reported in Section \ref{results} report the post episode side effect metric on a set of test environments different from the training environments used to train the agents.
\section{Method} \label{method}
\subsection{Training for Regularized Safe RL Agent}
\begin{figure}[h]
    \centering
    \includegraphics[width=\textwidth, height= 0.4\linewidth]{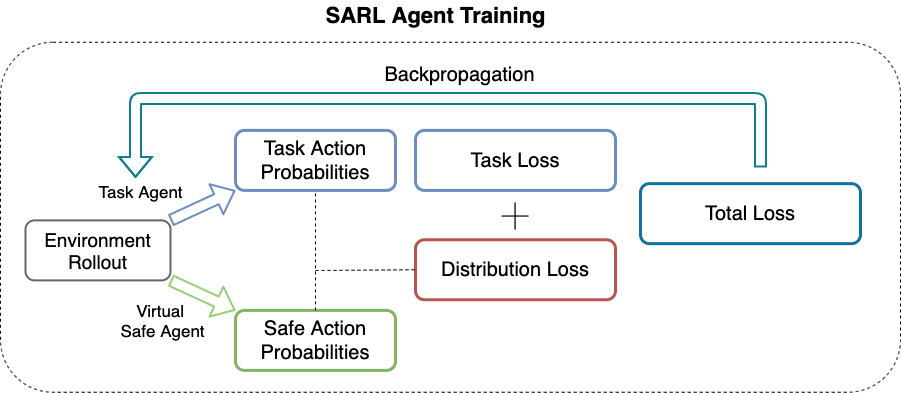}
    \caption{A co-training framework for safety aware RL training. The task agent,  \textcolor{dark-blue}{$A(\theta)$}, determines the action taken in the environment and the resulting trajectories. The virtual safe agent, \textcolor{dark-green}{$Z(\psi)$}, receives the same state as \textcolor{dark-blue}{$A(\theta)$} and makes a suggestion for a safe action given the state. The distance between the action probabilities of the \textcolor{dark-blue}{$A(\theta)$} and \textcolor{dark-green}{$Z(\psi)$} is captured in the Distribution Loss \textcolor{dark-red}{$\mathcal{L}_{\rm dist}(\theta, \psi)$}. \textcolor{dark-green}{$Z(\psi)$} learns how to maximize the safety objective on its own set of environments in parallel to \textcolor{dark-blue}{$A(\theta)$}.}
    \label{fig:co-training}
\end{figure}

Our method relies on regularizing the loss function of the RL agent with the distance of the task agent, \textcolor{dark-blue}{$A(\theta)$}, from the virtual safe actor, \textcolor{dark-green}{$Z(\psi)$}, as shown in Figure \ref{fig:co-training}.  

More formally, the general objective of the task agent \textcolor{dark-blue}{$A(\theta)$} can be expressed as:
\begin{align*}
    \mathcal{F_A}(\theta) &= \mathcal{L}_{\theta} +  \beta*\mathcal{L}_{\rm dist}(\mathbb{P}_{\pi_{\theta}}, \mathbb{P}_{\pi_{\psi}}) \tag{1} \label{eq:main_objective}
\end{align*}

where $\beta$ is a regularization hyperparameter, $\mathbb{P}_{\pi_{\theta}}$ represents the probability of taking a given action given by \textcolor{dark-blue}{$A(\theta)$}, and $\mathbb{P}_{\pi_{\psi}}$ represents the probability distribution of taking a given action according to \textcolor{dark-green}{$Z(\psi)$}. As shown in Equation \ref{eq:main_objective}, the actor loss $\mathcal{L}(\theta)$ is regularized by the distance between the actions suggested by the task agent and the virtual safe agent. The gradient of the objective in Equation \ref{eq:main_objective} expressed as the expectation of rewards of task agent actions $\alpha$ taken from a distribution of policies $\mathbb{P}_{A_{\theta}}$ is then given by:
\begin{align*}
    \nabla_{\theta}\mathcal{F_A}(\theta) &= \nabla_{\theta} \mathbb{E}_{\alpha \sim \mathbb{P}_{A_{\theta}}}[\mathcal{L}(\theta) ]  + \beta * \nabla_{\theta} \mathbb{E}_{\alpha \sim \mathbb{P}_{A_{\theta}}, \zeta \sim \mathbb{P}_{Z_{\psi}}} [\mathcal{L}_{\rm dist}( \alpha, \zeta)] \tag{2} \label{eq:grad_main_objective} \\
\end{align*}
where $\nabla_{\theta}$ is independent from the virtual safe agent actions \textcolor{dark-green}{$\zeta$} given that \textcolor{dark-green}{$Z(\psi)$} is only dependent on \textcolor{dark-green}{$\psi$}. This formulation enables training \textcolor{dark-green}{$Z(\psi)$} independently from \textcolor{dark-blue}{$A(\theta)$}, thereby abstracting the notion of safety away from the task.
The gradient formulation underscores the importance for a distance metric \textcolor{dark-red}{$\mathcal{L}_{\rm dist}$} that is differentiable to ensure that gradients update the task agent parameters \textcolor{dark-blue}{$\theta$} from both terms of the augmented loss functions.

\subsection{Distance Metrics for Loss Regularization}
The primary objective of the regularization term is to express a notion of distance between a purely reward based action and a purely safety motivated action, thereby penalizing \textcolor{dark-blue}{$A(\theta)$} for taking a purely reward motivated action. We model the regularization term as the distance between probability distributions $\mathbb{P}_{\pi_{\theta}}$ and $\mathbb{P}_{\pi_{\psi}}$ corresponding to  \textcolor{dark-blue}{$A(s|\theta)$} and \textcolor{dark-green}{$Z(s|\psi)$} respectively. The distance formulation between $\mathbb{P}_{\pi_{\theta}}$ and $\mathbb{P}_{\pi_{\psi}}$ intuitively captures how far the behavior of \textcolor{dark-blue}{$A(s|\theta)$} differs from \textcolor{dark-green}{$Z(s|\psi)$}.


Given this formulation, previous work \citep{nowozin2016f, Arjovsky2017, Huszar2015} has provided a number of choices for distance metrics in supervised learning problems with various advantages or shortfalls. One common method of measuring the difference in probability distributions is the KL Divergence,  $D_{KL}(p \| q) = \int_x p(x) \log \frac{p(x)}{q(x)} dx$, where $p$ and $q$ are probability distributions described by probability density functions. 

The KL Divergence, however, has some significant disadvantages -- the most significant one being that the KL Divergence is unbounded when probability density functions to express the underlying distribution cannot be easily described by the model manifold \citep{Arjovsky2017}. Furthermore, the KL divergence is not symmetric given that $D_{KL}(p \| q) \neq D_{KL}(q \| p)$, and also does not satisfy the triangle inequality. One alternative to the KL Divergence is the Jensen-Shannon distance $D_{JS}(p \| q) = \frac{1}{2} D_{KL}(p \| m)+ \frac{1}{2} D_{KL}(q \| m )$ with $m = \frac{1}{2}(p+q)$, which is symmetric, satisfies the triangle inequality and is bounded: $0 \leq D_{JS} \leq \log(2)$. These advantages make $D_{JS}$ a good choice for the SARL algorithm, but as discussed extensively in \citet{Arjovsky2017}, $D_{JS}$ also has notable disadvantages, the most important being that  $D_{JS}$ is not guaranteed to always be continuous and differentiable in low manifold settings.

Another alternative to $D_{JS}$ is the Wasserstein Distance. As discussed in \citet{Arjovsky2017}, the Wasserstein Distance is generally better suited for calculating distances for low-dimensional manifolds compared to $D_{JS}$ and other variants of the KL divergence. In its analytical form the Wasserstein Distance $W_p(P, Q) = 
(\inf_{J \in \mathcal{J}(P,Q)} \int \lVert x-y \rVert^p \mathrm{d}J(x,y))^{\frac{1}{p}}$, however, is intractable to compute in most cases leading many researchers to establish approximations of the metric. A common way of approximating the Wasserstein Distance is to re-formulate the calculation as an optimal transport problem of moving probability mass from $p$ to $q$, as shown in \citet{Cuturi2013} and \citet{Pacchiano2019}. The dual formulation based on behavior embedding maps of policy characteristics described in \citet{Pacchiano2019} is particularly applicable for the SARL algorithm, leading us to adapt it as an additional alternative to the Jensen-Shannon Distance. In this formulation, policy characteristics are converted to distributions in a latent space of behavioral embeddings on which the Wasserstein Distance is then computed.

For our experiments in Section \ref{experiments}, we apply both $D_{JS}$ and the dual formulation of the Wasserstein Distance described in \citet{Pacchiano2019} to regularize between \textcolor{dark-blue}{$A(s|\theta)$} and \textcolor{dark-green}{$Z(s|\psi)$}. 

\subsection{Safety Aware Reinforcement Learning}
The paragraphs above in Section \ref{method} describe the individual components of SARL. The experiments outlined in Section \ref{experiments} discuss SARL applied to Proximal Policy Optimization (PPO) \citep{schulman2017}. \citet{Wainwright2020} applied PPO for solving the different environments in the SafeLife suite, making SARL-PPO a natural extension. The loss formulation $\mathcal{L}_{\theta}^{PPO}$ used in Algorithm \ref{alg:co-train-algo} is the same as the one decribed in \citet{schulman2017}:  
\begin{align*}
    \mathcal{L}_{\theta}^{PPO} &= \mathbb{E}_t[\mathcal{L}_{t}^{Clip}(\theta) -  c_1\mathcal{L}_{t}^{Value}(\theta) +  c_2S[\pi_{\theta}](s_t)]
    \tag{3} \label{eq:ppo_loss}
\end{align*}
As shown in more detail in Algorithm \ref{alg:co-train-algo}, \textcolor{dark-blue}{$A(\theta)$} is trained using the regularized loss objective described in Equation \ref{eq:main_objective}, while \textcolor{dark-green}{$Z(\psi)$} is trained exclusively on $\mathcal{L}_{\theta}^{PPO}$ using the frame-by-frame side effect information as the reward.

\vspace{\baselineskip}
\begin{algorithm}[]
\caption{SARL - PPO}
\label{alg:co-train-algo} 
\begin{algorithmic}[1]
\State Initialize an actor \textcolor{dark-blue}{$A(s|\theta)$} and virtual safety agent \textcolor{dark-green}{$Z(s|\psi)$}
\State Set hyperparameters for \textcolor{dark-blue}{$A(s|\theta)$}, \textcolor{dark-green}{$Z(s|\psi)$} and distance metric \textcolor{dark-red}{$\mathcal{L}_{\rm dist}$}
    
    \While{training SARL-PPO}
        \For{each actor update \textcolor{dark-blue}{$A(\phi)$}}
        \State Run \textcolor{dark-blue}{$A(\phi)$}  to generate a minibatch of transitions \textcolor{dark-blue}{$\alpha$} with task rewards \textcolor{dark-blue}{$r$}
        \State Run \textcolor{dark-green}{$Z(\psi)$} to generate a minibatch of transitions \textcolor{dark-green}{$\zeta$} 
        \State  Compute \textcolor{dark-blue}{$\mathcal{L}_{\theta}^{PPO}$} and \textcolor{dark-blue}{$\mathbb{P}_{A_{\theta}}$} using transitions in \textcolor{dark-blue}{$\alpha$} 
        \State  Compute  \textcolor{dark-green}{$\mathbb{P}_{Z_{\psi}}$} using transitions in \textcolor{dark-green}{$\zeta$} 
        \State Optimize \textcolor{dark-blue}{$A(\theta)$} using $\textcolor{dark-blue}{\mathcal{L}_{\theta}^{PPO}} +  \textcolor{dark-red}{\beta*\mathcal{L}_{\rm dist}(}\textcolor{dark-blue}{\mathbb{P}_{A_{\theta}}}, \textcolor{dark-green}{\mathbb{P}_{Z_{\psi}}}\textcolor{dark-red}{)}$
         \EndFor
        
        \For{each virtual agent update \textcolor{dark-green}{$Z(\psi)$}}
        \State Run \textcolor{dark-green}{$Z(\psi)$} to generate a minibatch of transitions \textcolor{dark-green}{$\zeta$} with safety metric \textcolor{dark-green}{$s$}
        \State  Compute \textcolor{dark-green}{$\mathcal{L}_{\psi}^{PPO}$} using transitions in \textcolor{dark-green}{$\zeta$}
        \State Optimize \textcolor{dark-green}{$Z(\psi)$} using $\textcolor{dark-green}{\mathcal{L}_{\psi}^{PPO}}$ with \textcolor{dark-green}{$s$} as the reward
        \EndFor

    \EndWhile
\end{algorithmic}
\end{algorithm}

The training algorithm is agnostic to the side effect metric, \textcolor{dark-green}{$s$}, used in the environment, leading itself to a plug-and-play approach where the virtual safe agent can modulate the task agent for a variety of different environment specific side effect metrics without major modification to the overall structure of the method.

In addition to training both \textcolor{dark-blue}{$A(s|\theta)$} and \textcolor{dark-green}{$Z(s|\psi)$} from scratch as shown in Algorithm \ref{alg:co-train-algo}, we also perform zero-shot generalization of a previously trained \textcolor{dark-green}{$Z(s|\psi)$} to investigate whether the concept of side effects can be abstracted out of the environmental dynamics and the intricacies of the task. In this case, lines 13-17 from Algorithm \ref{alg:co-train-algo} are not performed as no updates for \textcolor{dark-green}{$Z(s|\psi)$} are required, with \textcolor{dark-green}{$Z(s|\psi)$} only being used to modulate the behavior of \textcolor{dark-blue}{$A(s|\theta)$} via the distance metric regularization. 

\subsection{Tracking the Champion Policy}
The SafeLife suite includes a complex set of procedurally generated environments, which can lead to a significant amount of variability throughout training and testing episodes. In order to account for this variability, we track the best policy throughout the training process for a fixed set of test levels for the different metrics we care about, specifically episode length, performance ratio and side effects, as described in Algorithm \ref{alg:champion-policy-tracking}. 

\vspace{\baselineskip}
\begin{algorithm}
\caption{Champion Policy Tracking}
\label{alg:champion-policy-tracking} 
\begin{algorithmic}[1]

\State Initialize Training and Champion Policy \textcolor{dark-red}{$C(\theta)$}
\For{Every $k$ Environment Steps}
    \State Evaluate task agent \textcolor{dark-blue}{$A(\theta)^k$} on fixed set of test levels
    \If{\textcolor{dark-blue}{$Score_{A(\theta)^k}$} $>$ \textcolor{dark-red}{$Score_{C(\theta)}$}}
        \State \textcolor{dark-red}{$C(\theta)$} $=$ \textcolor{dark-blue}{$A(\theta)^k$}
    \EndIf
\EndFor

\end{algorithmic}
\end{algorithm}
The champion policies operate on test levels, where no learning occurs, and track test-level metrics. This is particularly relevant to the side effect metric, described in \Cref{safelife-env}, where we track the episodic side effect even though training occurs with frame-by-frame impact measure. We describe these metrics in greater detail in Section \ref{experiments}.
\section{Experiments} \label{experiments}
Our experiments contain the following algorithmic runs:
\begin{itemize}
    \item The reward-penalty baseline method described in \citep{Wainwright2020} where the impact penalty of a given action is subtracted from the reward the agent receives for that particular frame.
    \item SARL agents where both the actor \textcolor{dark-blue}{$A(\theta)$} and virtual safety agent \textcolor{dark-green}{$Z(\psi)$} are training from scratch using the Jensen-Shannon Distance as well as the dual formulation of the Wasserstein Distance described in \citep{Pacchiano2019}
    \item SARL agents where \textcolor{dark-blue}{$A(\theta)$} is trained  while \textcolor{dark-green}{$Z(\psi)$} is taken zero-shot from a previous training run. The main purpose of this experiment is to show that the concept of side-effects in the SafeLife suite can be abstracted from the specific task (prune vs append) and the specific environment setting (still vs dynamic). The ability to extract a notion of side effects that does not rely on environmental signal for every frame enables us to train the virtual safety agent \textcolor{dark-green}{$Z(\psi)$} only once, usually on the simplest task, which can then be used to influence any agent on any subsequent task.
\end{itemize}
We conduct our experiments on four different tasks in the SafeLife suite: prune-still, append-still, prune-dynamic, append-dynamic. As described in Section \ref{safelife-env}, dynamic environments have natural variation independent of the actions of the agents, while all changes in still environments can be traced back to the actions of the agent. We evaluate our champion policies \textcolor{dark-red}{$C(\theta)$} every 100,000 environment steps on the episode length across a set of 100 different testing environments whose configurations are not part of the configurations of environments used in the training process. The length of an episode is the number of steps the agent takes to complete an episode, where a shorter length indicates that the agent can solve the task better and more efficiently. In our results in Section \ref{results} we show the standard error of the champion measured in performance given by the ratio $\frac{\rm agent \: reward}{\rm possible \: reward}$ and the cumulative side effect measure described in Section \ref{safelife-env} and \citet{Wainwright2020}.

The experimental results have a strong dependency on hyperparameters chosen, specifically the impact penalty fraction in the reward penalty baseline baseline and the regularization parameter $\beta$ in SARL. Changing these parameters generally results in non-linear trade-offs between episode length, performance and side effects, meaning that policies with high performance often have high side effects and policies with low side effects often have low performance. In the cases of low side effects and low performance, the agent does not perform any significant actions that would either negatively (side effect) or positively (reward) disturb the environment. Our ideal goal is to have a policy that is both performant on the task and has low side effects. As such, in \Cref{results} we describe results of experiments that in our best judgement represent the best cases of such policies, and apply the same regularization hyperparameters across all environments. The full set of our algorithmic and regularization hyperparameters are shown in \Cref{sec:hyperparameters}. As discussed in more detail in \Cref{discussion}, for future work we aim, and encourage others, to obtain Pareto optimal frontiers that describe the trade-off for the regularization hyperparameters more thoroughly. 

\section{Results} \label{results}

\begin{figure}[ht]
\subfigure[Prune-Still Length]
{\includegraphics[width=0.33\columnwidth,height=0.25\linewidth]{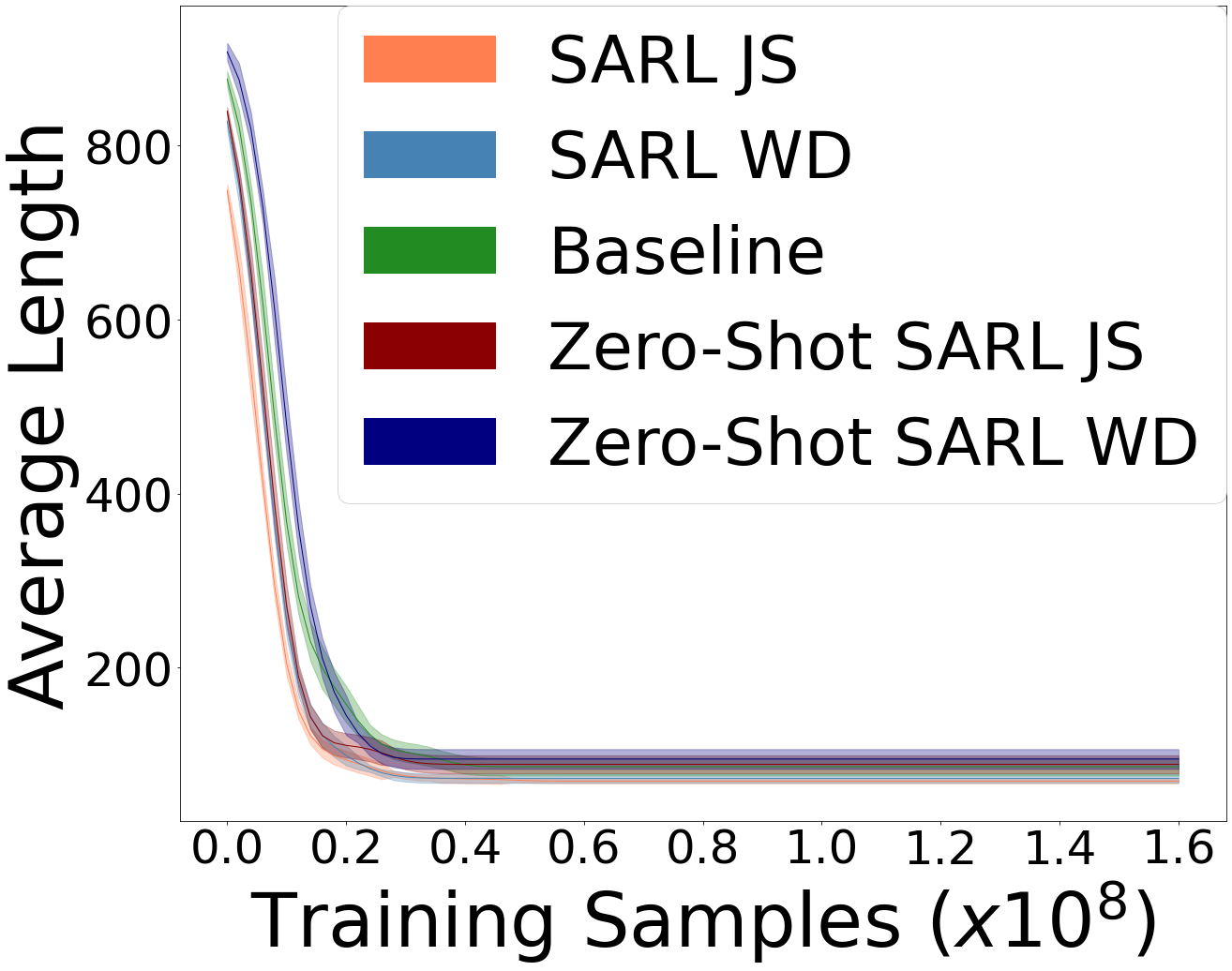}}\hspace*{\fill}
\subfigure[Prune-Still Performance]
{\includegraphics[width=0.33\columnwidth,height=0.25\linewidth]{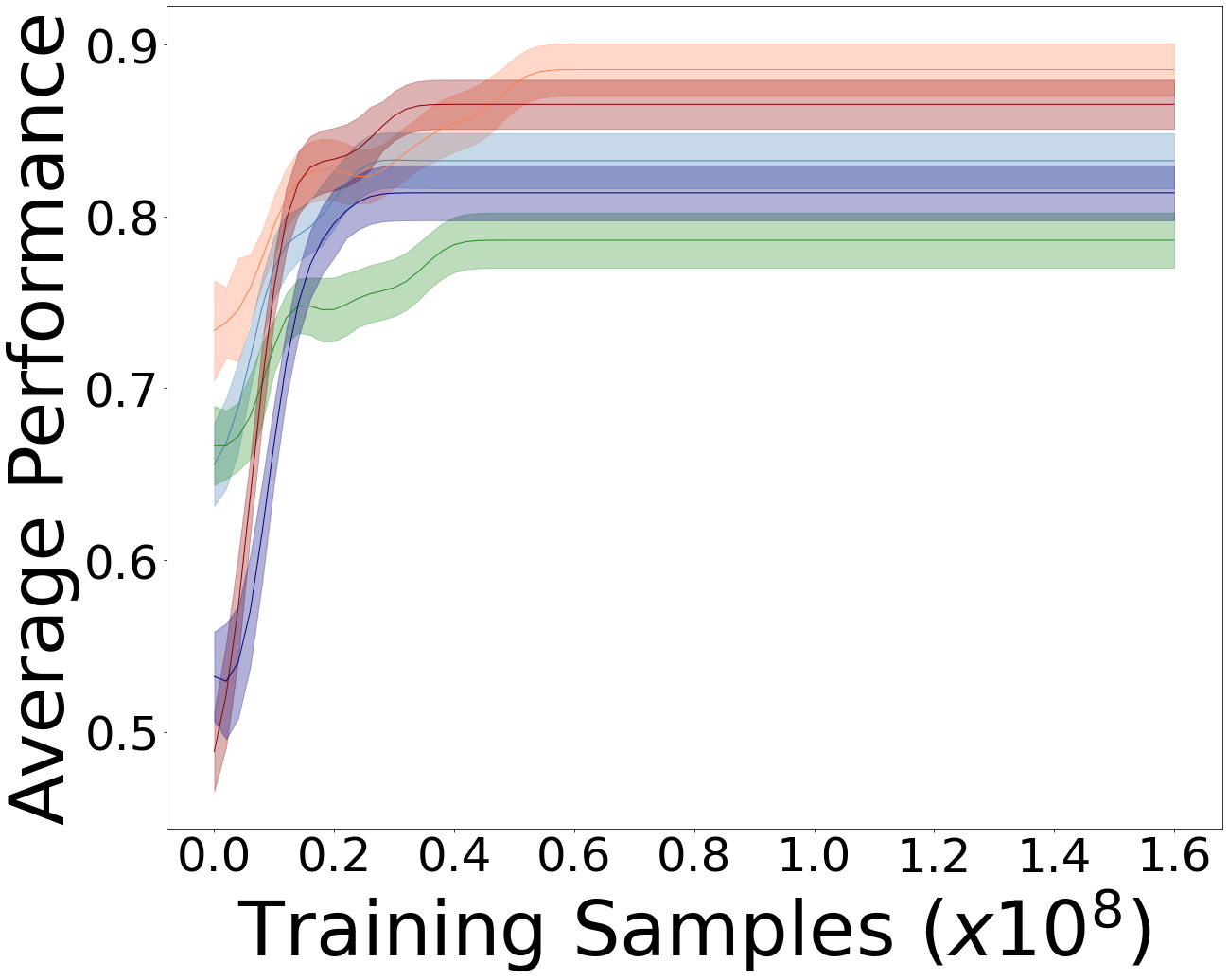}}\hspace*{\fill}
\subfigure[Prune-Still Side Effects]
{\includegraphics[width=0.33\columnwidth,height=0.25\linewidth]{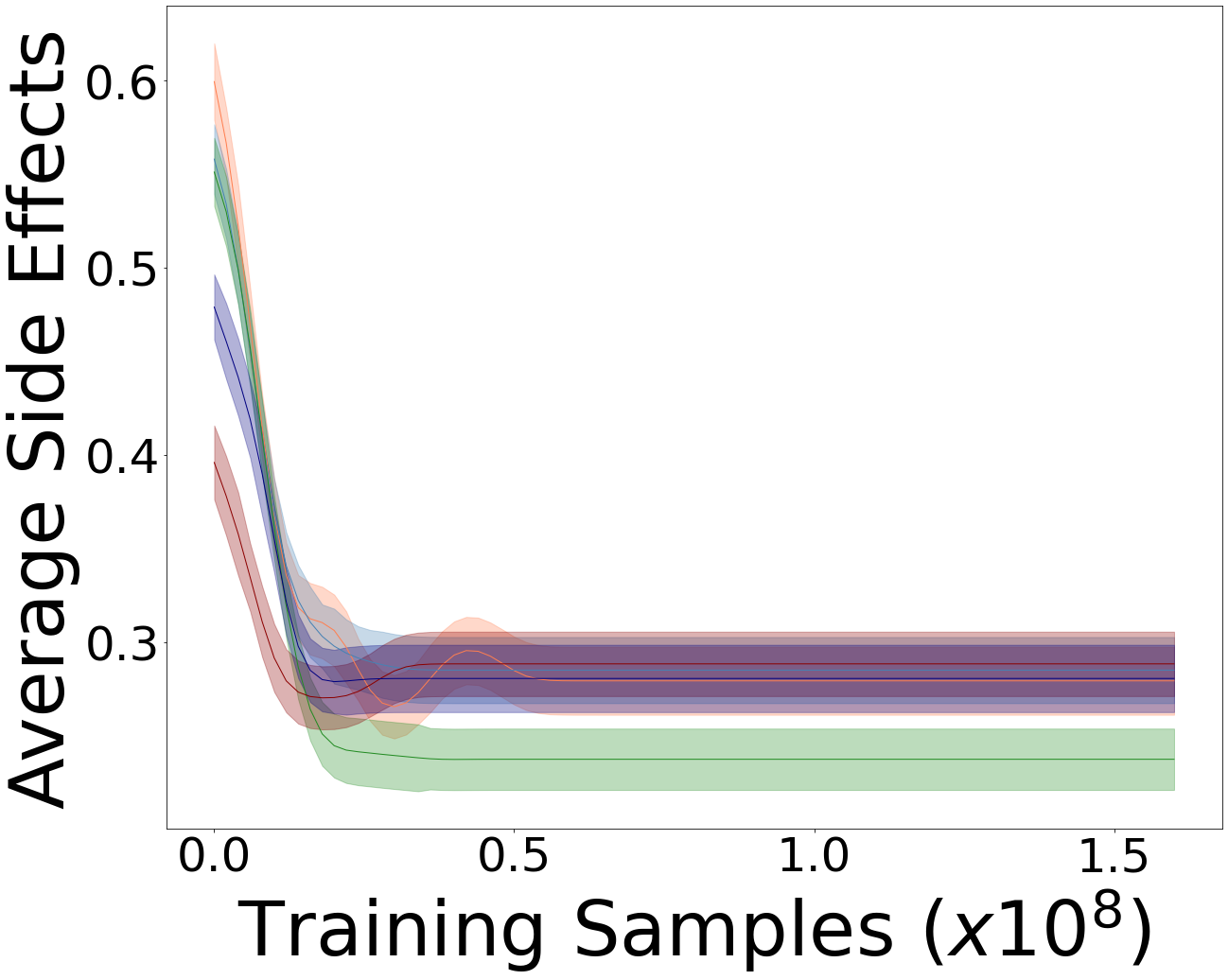}}\hspace*{\fill}

\subfigure[Append-Still Length]
{\includegraphics[width=0.33\columnwidth,height=0.25\linewidth]{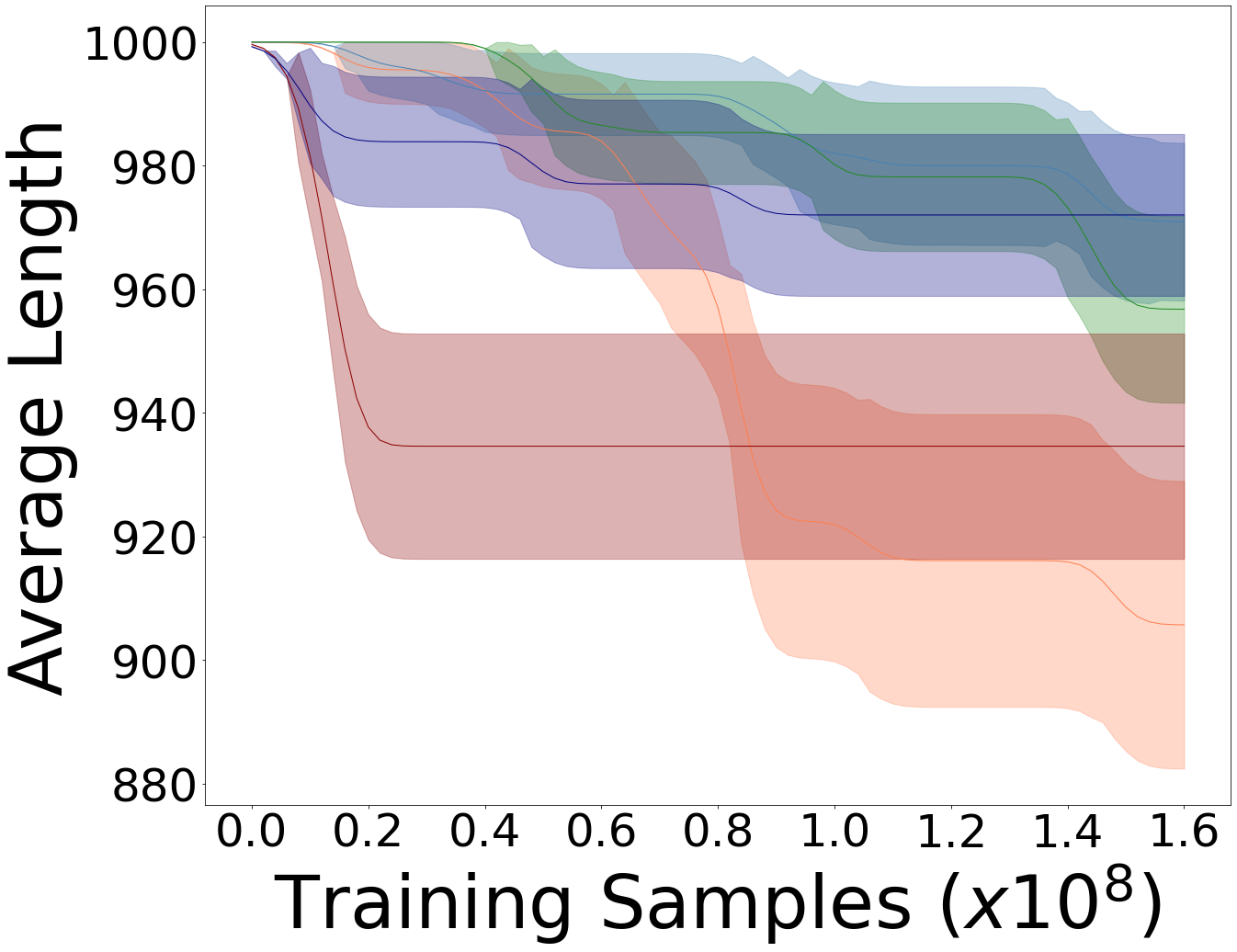}}\hspace*{\fill}
\subfigure[Append-Still Performance]
{\includegraphics[width=0.33\columnwidth,height=0.25\linewidth]{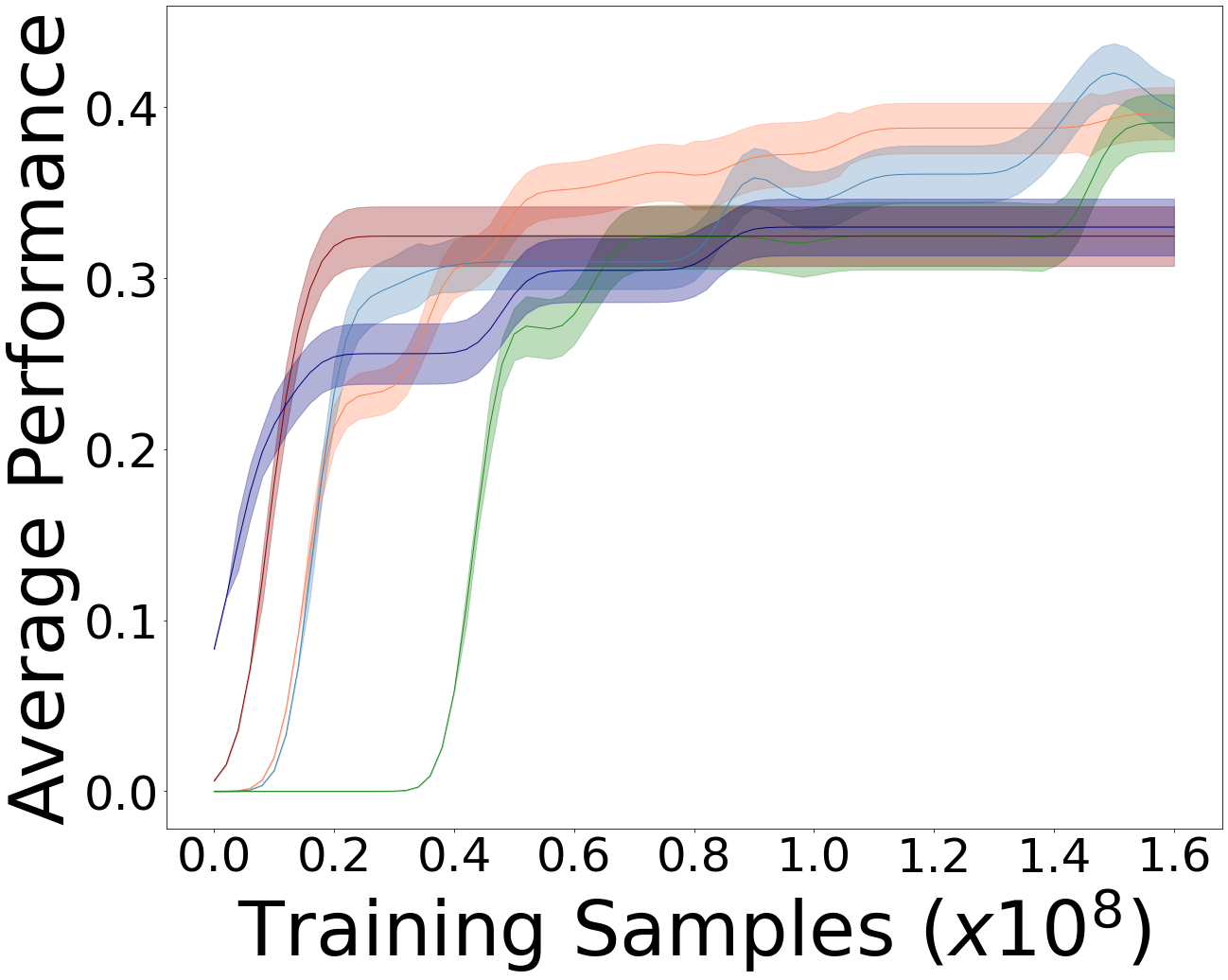}}\hspace*{\fill}
\subfigure[Append-Still Side Effects]
{\includegraphics[width=0.33\columnwidth,height=0.25\linewidth]{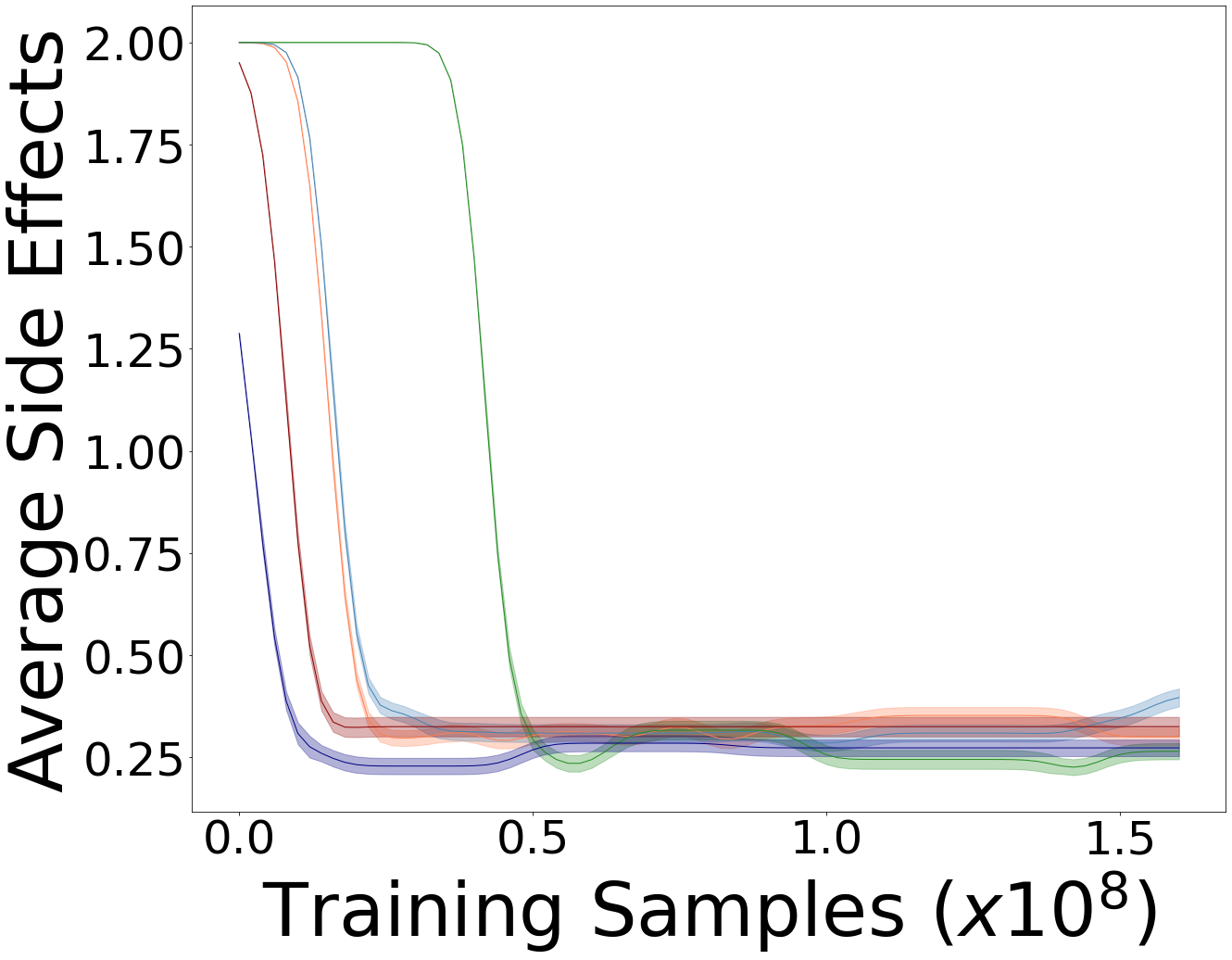}}\hspace*{\fill}

\subfigure[Prune-Dynamic Length]
{\includegraphics[width=0.33\columnwidth,height=0.25\linewidth]{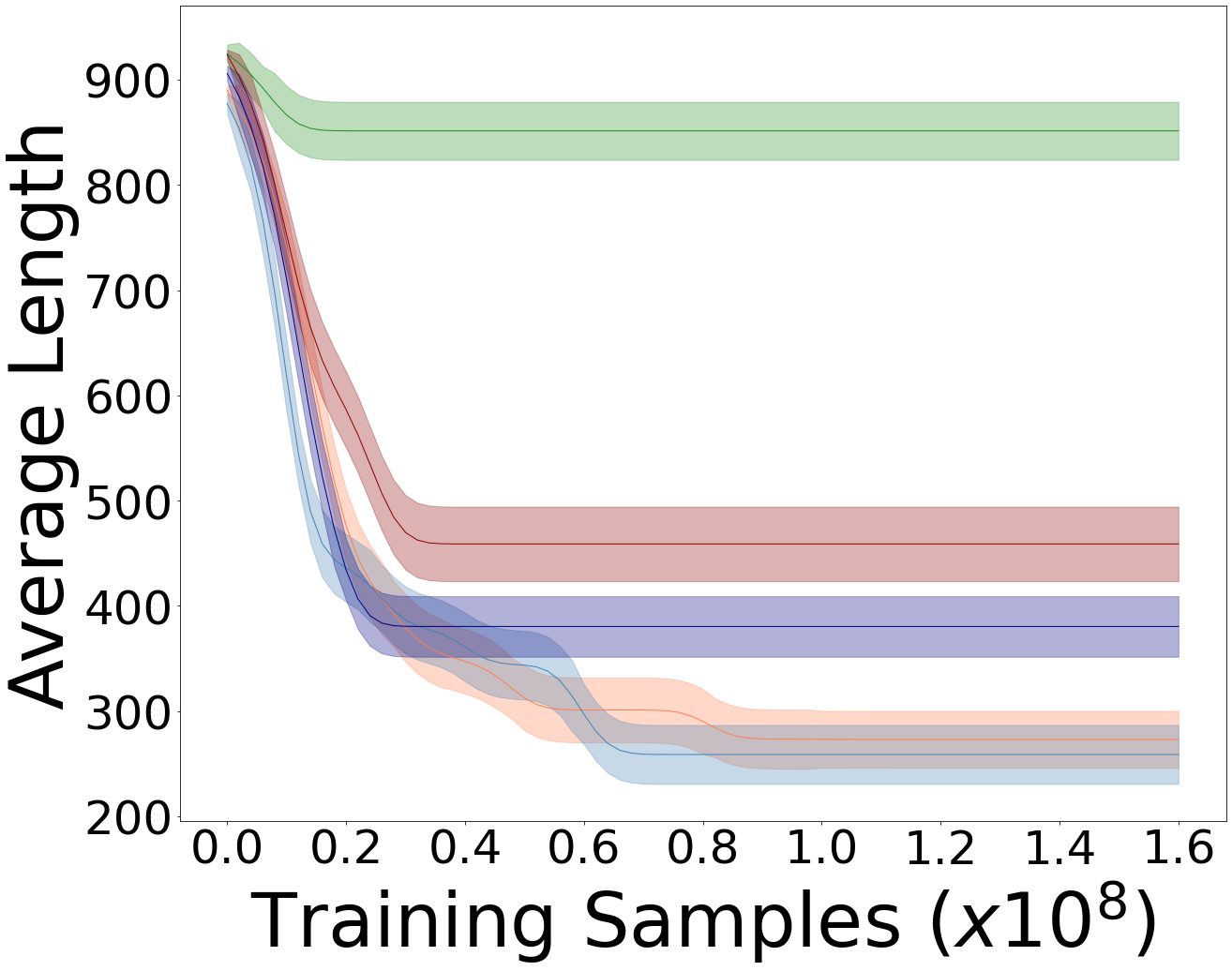}}\hspace*{\fill}
\subfigure[Prune-Dynamic Performance]
{\includegraphics[width=0.33\columnwidth,height=0.25\linewidth]{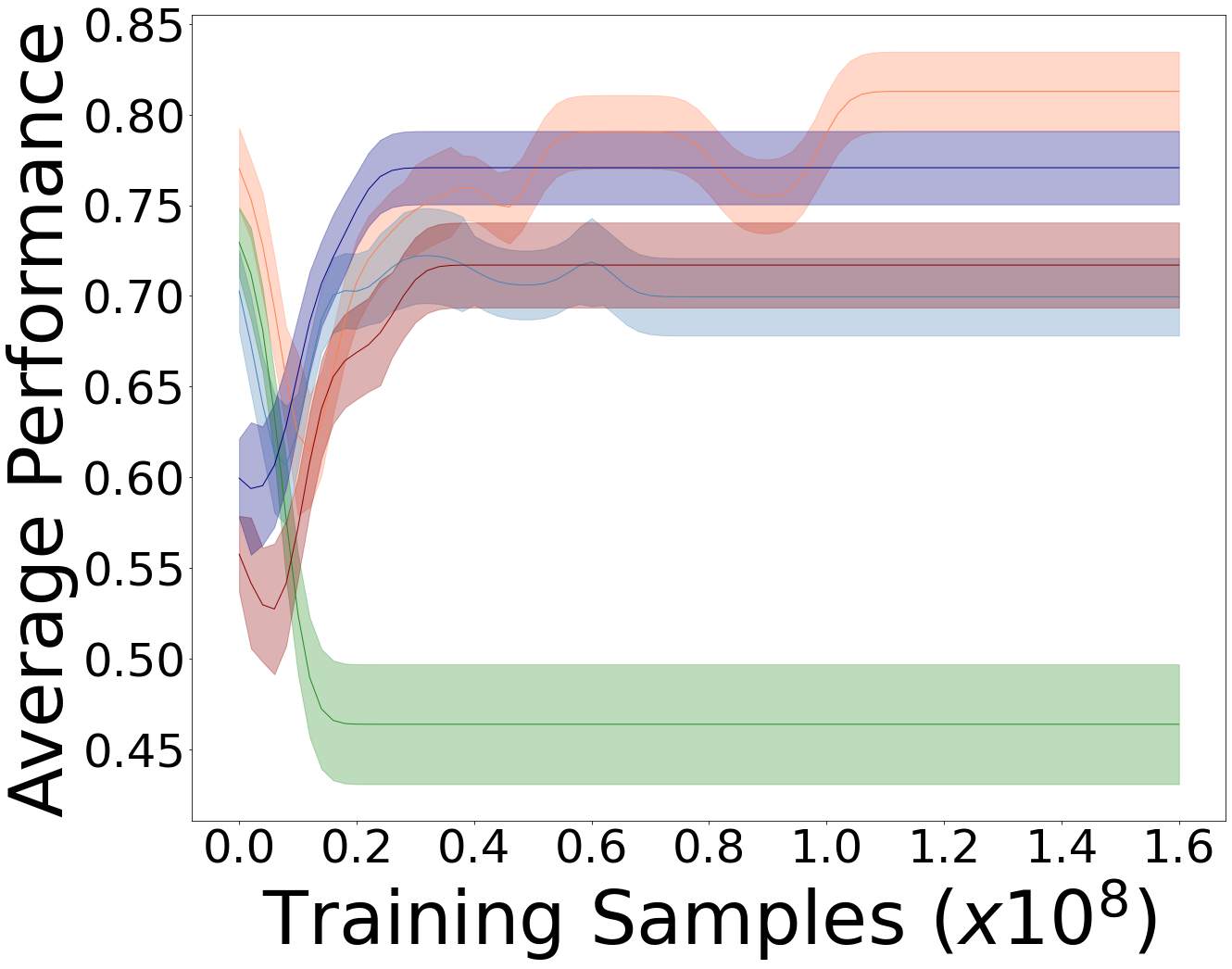}}\hspace*{\fill}
\subfigure[Prune-Dynamic Side Effects]
{\includegraphics[width=0.33\columnwidth,height=0.25\linewidth]{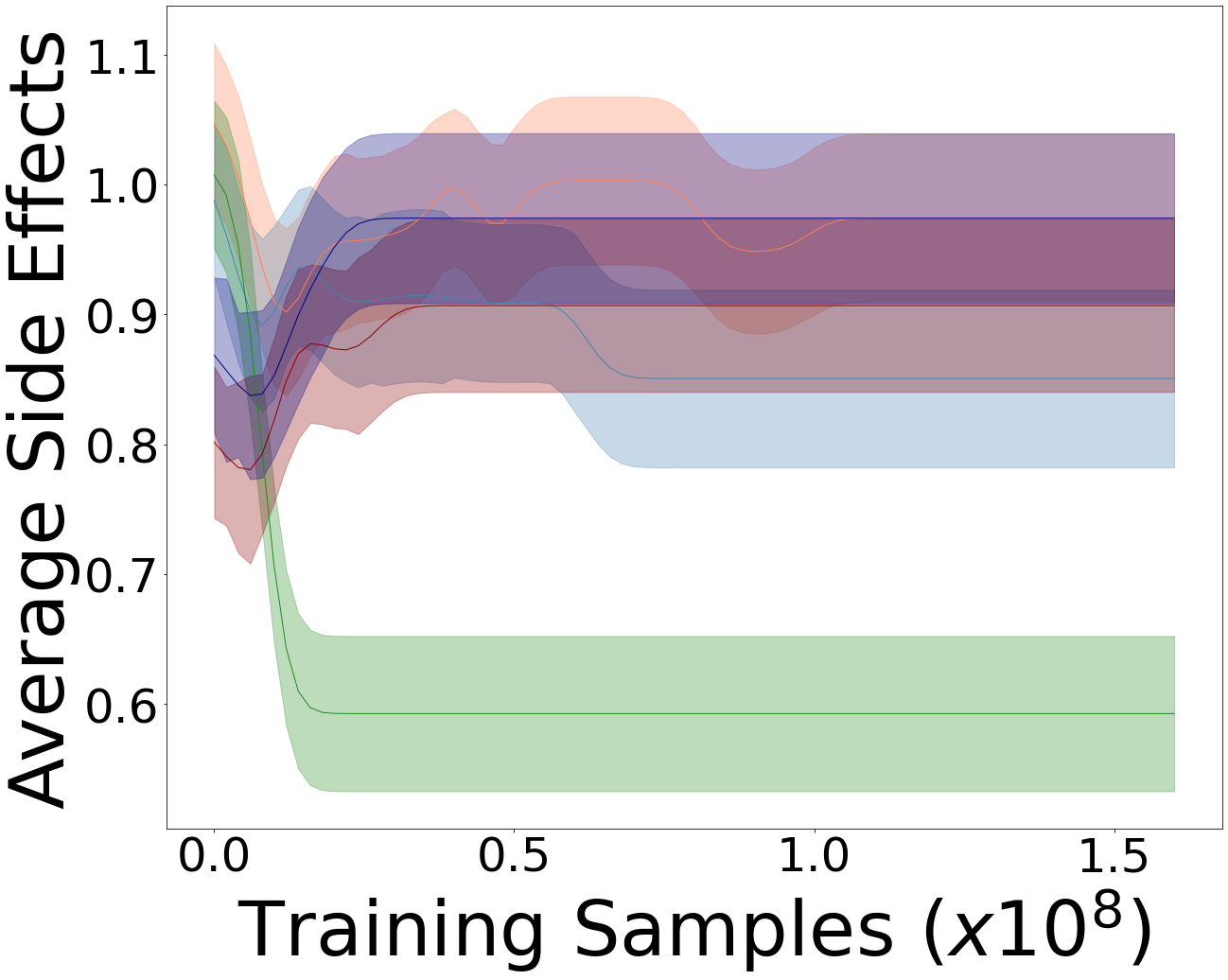}}\hspace*{\fill}

\subfigure[Append-Dynamic Length]
{\includegraphics[width=0.33\columnwidth,height=0.25\linewidth]{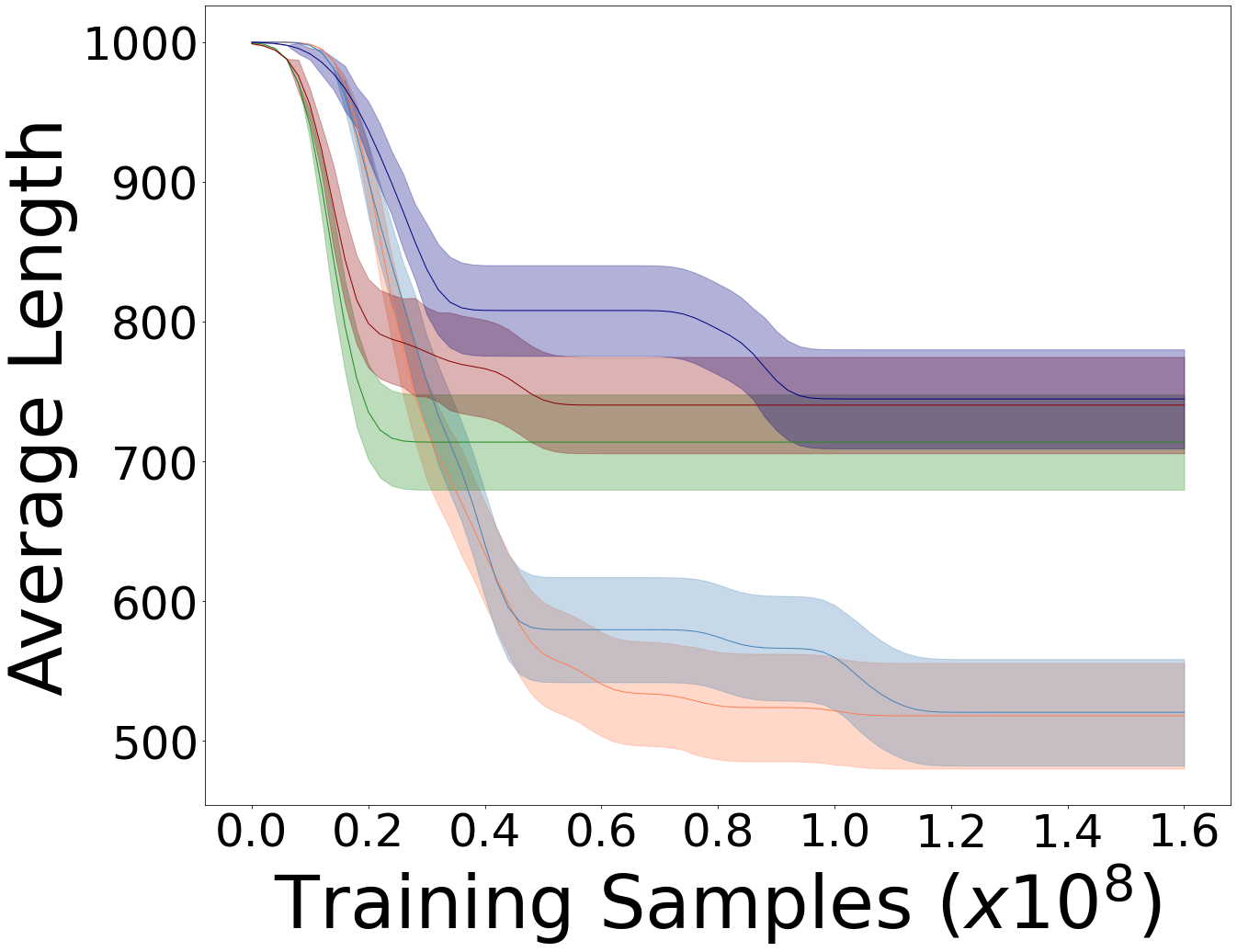}}\hspace*{\fill}
\subfigure[Append-Dynamic Performance]
{\includegraphics[width=0.33\columnwidth,height=0.25\linewidth]{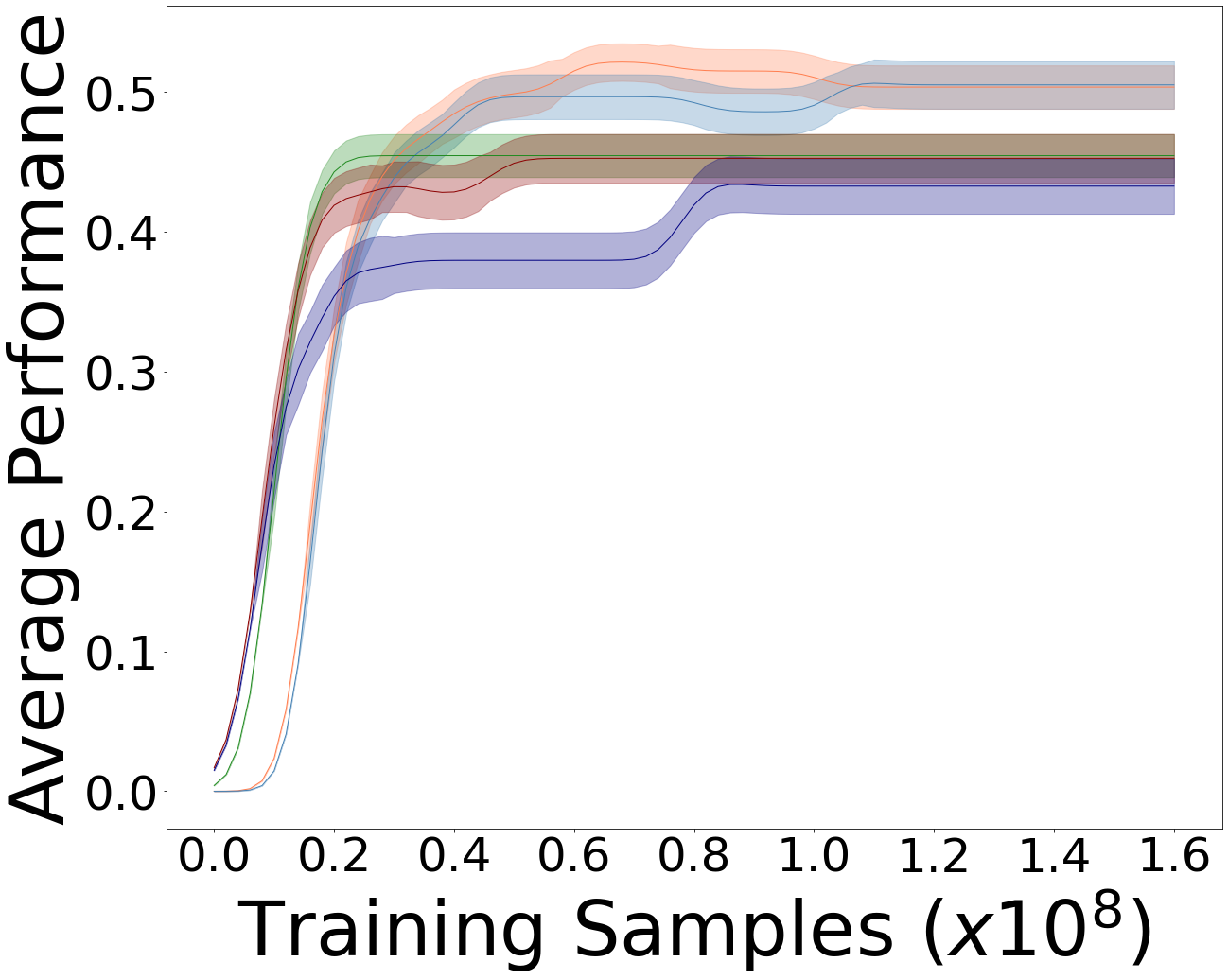}}\hspace*{\fill}
\subfigure[Append-Dynamic Side Effects]
{\includegraphics[width=0.33\columnwidth,height=0.25\linewidth]{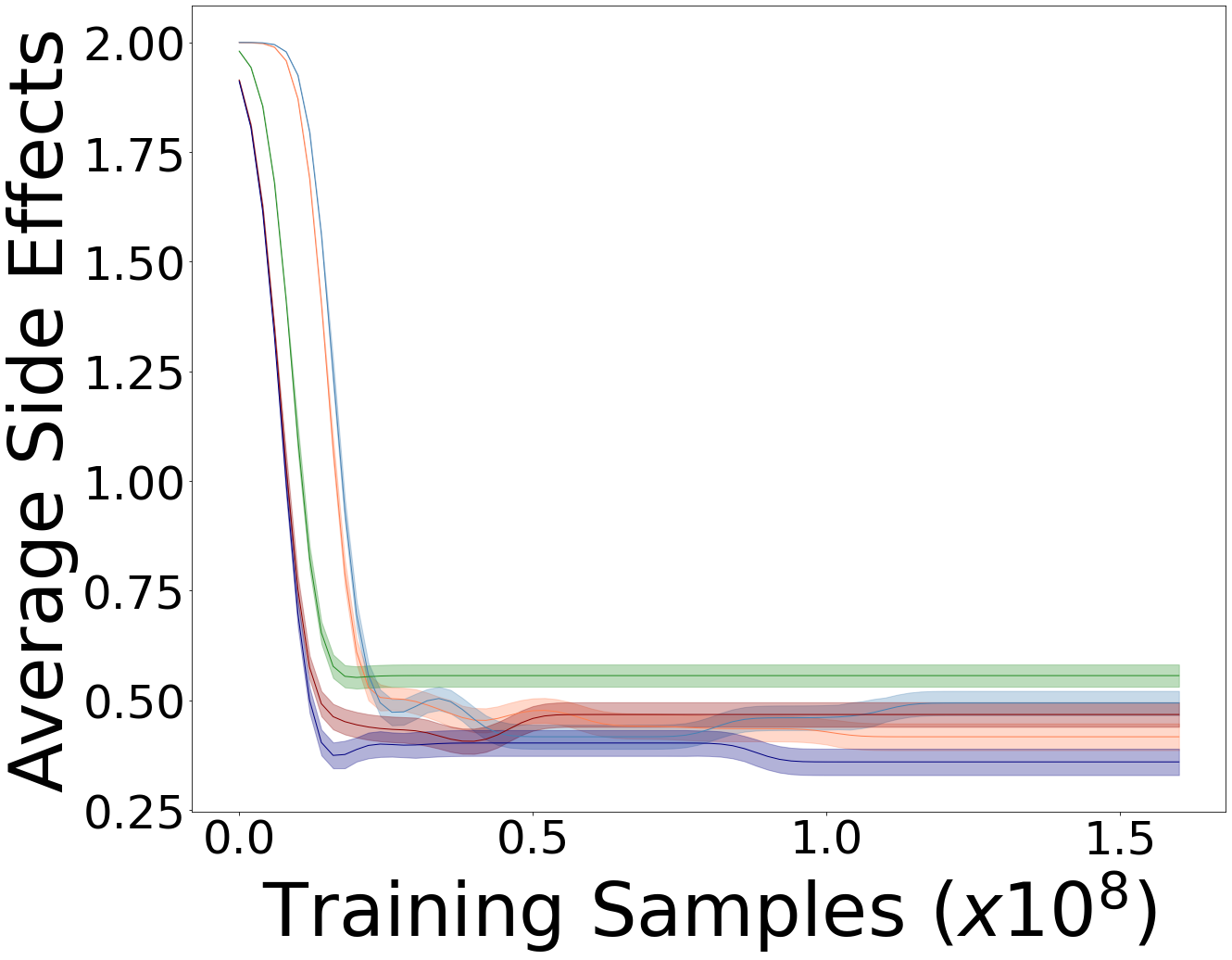}}\hspace*{\fill}

\caption{Length Champion for SafeLife Suite of \emph{1. prune-still (a-c), 2. append-still (d-f), 3. prune-dynamic (g-i), 4. append-dynamic (j-l)} tasks evaluated for 100 testing environments every 100,000 steps on \emph{Episode Length (left column)} where shorter is better, \emph{Performance Ratio (middle column)} where higher is better, and \emph{Episodic Side Effect (right column)} where lower is better}
\label{fig:length-champ-results}
\end{figure}

The results of the experiments shown in \Cref{fig:length-champ-results} demonstrate that a virtual safety agent trained on one task in the SARL framework can generalize zero-shot to other tasks and environment settings in the SafeLife suite, while maintaining competitive task and side effect scores compared to the baseline method. This allows us to abstract the notion of safety away from the environment specific side effect metric, and also increase the overall sample efficiency of the SARL method for subsequent training runs. The SARL methods that are trained from scratch also show competitive task and side effect scores compared to the baseline method.

In the still environment we chose to generalize the virtual safety agents from \emph{prune-still} and \emph{append-still} to the other tasks using both distance metrics. The results show that zero-shot generalization of \textcolor{dark-green}{$Z(\psi)$} matches the behavior of SARL trained from scratch, as well as matching or outperforming the baseline method on episode length and performance. 

\textbf{Prune-Still Environment:} The reward penalty baseline matches the episode length of all other methods while maintaining a slightly lower performance and side effects. All SARL methods, including both metrics and zero-shot SARL, generally perform equally well on length and side effects while the $SARL_{D_{JS}}$ has a slightly better performance than $SARL_{D_{WD}}$.

\textbf{Append-Still Environment:} The reward penalty baseline generally matches the performance and side effects of the SARL methods, while slightly underperforming $SARL_{D_{JS}}$ on episode length. $SARL_{D_{JS}}$ generally performs better on episode length compared to the other methods, both in training from scratch and the zero-shot experiments. 

For the dynamic environment we chose to generalize the virtual safety agents trained on \emph{prune-still} and \emph{append-still} to the dynamic environment tasks using both distance metrics. In the zero-shot experiments, we applied the version of \textcolor{dark-green}{$Z(\psi)$} that is furthest away from the given setting, meaning \emph{prune-still} is generalized to \emph{append-dynamic} and \emph{append-still} is generalized to \emph{prune-dynamic}. The results show that zero-shot generalization of \textcolor{dark-green}{$Z(\psi)$} matches the behavior of SARL trained from scratch, as well as matching or outperforming the baseline method on some metrics. 

\textbf{Prune-Dynamic Environment:} In this environment, we observe that the baseline method cannot solve the task, as shown by the fact that the episode length does not decrease significantly. However, it incurs very little side-effect cost.  This indicates that the baseline agent is acting safer by not doing much in the environment, but actually fails to solve the primary task. All SARL agents outperform the baseline on episode length and performance ratio, indicating that SARL effectively learns the task. 

\textbf{Append-Dynamic Environment:} The reward penalty baseline generally matches the behavior of the zero-shot SARL methods on episode length, performance and side effects. The SARL method trained from scratch, both $SARL_{D_{JS}}$ and $SARL_{D_{WD}}$ outperform the baseline as well as the zero-shot experiments on episode length and slightly on performance

\section{Discussion} \label{discussion}
In this work, we explored the prospect of regularizing the loss function of an RL agent using distance metrics that encapsulate a notion of safe behavior for the RL agent. 
We believe this work shows the promise of this approach to train RL agents in environments where side effects are important. As mentioned in \Cref{introduction}, side effects are often difficult to define, especially when interwoven with the primary task, and therefore measuring and interpreting side effects is an ongoing area of research. In order for our framework to be easily adoptable, we designed it to be flexible to different side effect metrics. \par

The idea of using suitable distance metrics to perform co-training of multiple RL agents has a variety of future research directions. One such avenue is the development of new distance metrics, including different variations of the Wasserstein Distance, as well as ones that can exploit various channels of information that we did not consider in out work \citep{parker2020effective}. The ideal distance metric would capture both the information richness from the different channels and encode a notion of a safety objective which can then be transferred to the primary agent to influence its behavior. \par

There also exists a great opportunity to apply techniques from multi-objective optimization to side effect problem. The literature is rich with multi-objective optimization problems in supervised learning \citep{Ma2020} \citep{Sener2018} and reinforcement learning \citep{morl2019yang} \citep{Xu2020} that show promising approaches to adopt a robust multi-objective framework to the side effect problem. The greatest promise of a multi-objective framework is the possibility of obtaining Pareto fronts \citep{morl2019yang} that describe the optimal trade-off between task performance and safety in a given environment, which would be immensely valuable to making decisions in real-world environments. \par

Lastly, the training framework we proposed focused exclusively discrete action spaces which do not capture the full extent of RL algorithms and environments. As such, a natural extension of this work is to develop a framework for continuous action spaces that builds on the ideas presented here. Many continuous space algorithms have a variety of agents, such as actors and critics, working together to achieve a common objective, and we believe that integrating the idea of virtual agents with proper distance metrics can open up new algorithmic designs to tackle safety critical applications in reinforcement learning.


\clearpage
\bibliographystyle{abbrvnat}
\bibliography{safe_rl.bib}

\clearpage
\appendix \label{appendix}
\section{Implementation Details}
\label{sec:hyperparameters}
\begin{table}[h]

\begin{tabular}{|c||c|} 
 \hline
 Hyperparameter & Value\\ 
 \hline \hline
  $\gamma$ & $0.97$ \\ 
   Learning Rate & $3e^{-4}$ \\ 
    Batch Size  & $64$ \\ 
    Epochs per Training Batch   & $3$ \\ 
    Environment Steps per Training Iteration   & $20$ \\ 
    PPO Entropy Weight   &  $0.01$\\ 
    PPO Entropy Clip   &  $1.0$\\ 
    PPO Value Loss Coefficient   &  $0.5$\\ 
    PPO Value Loss Clip   &  $0.2$\\ 
    PPO Policy Loss Clip   &  $0.2$\\

 \hline \hline
\end{tabular}
\caption{Hyperparameters for PPO-SARL}
\label{varied}
\vspace{2em}
\end{table}

\begin{table}[h]

\begin{tabular}{|c||c||c||c||c||c|} 
 \hline
 Environment & Baseline & SARL JS & SARL WD & SARL Zero-Shot JS & SARL Zero-Shot WD\\ 
 \hline \hline
  Prune-Still & $0.3$ & $0.01$ & $0.01$ & $0.005$ & $0.005$ \\ 
  Append-Still & $0.3$ & $0.01$ & $0.01$ & $0.005$ & $0.005$ \\ 
  Prune-Dynamic & $0.3$ & $0.01$ & $0.01$ & $0.005$ & $0.005$ \\ 
  Append-Dynamic & $0.3$ & $0.01$ & $0.01$ & $0.005$ & $0.005$ \\

 \hline \hline
\end{tabular}
\caption{Hyperparameters for Side Effect Procedures -- Baseline: Impact Penalty; SARL: $\beta$}
\label{varied}
\vspace{2em}
\end{table}

\end{document}